%% file: root.tex
\documentclass[letterpaper, 10 pt, conference]{ieeeconf}  

\IEEEoverridecommandlockouts                              

\usepackage{kotex}

\usepackage{graphicx}
\usepackage{multirow}
\usepackage{color}
\usepackage[hidelinks]{hyperref}
\usepackage{amsmath}
\usepackage{booktabs}
\usepackage{adjustbox}
\usepackage{makecell} 
\usepackage{subcaption} 
\usepackage[font=small]{caption}
\usepackage{pifont} 
\usepackage{amssymb} 
\usepackage[symbol]{footmisc} 
\usepackage{tablefootnote}
\hypersetup{
    colorlinks=true,
    linkcolor=black,
    filecolor=black,      
    urlcolor=magenta,
}
\makeatletter
\def\endthebibliography{%
  \def\@noitemerr{\@latex@warning{Empty `thebibliography' environment}}%
  \endlist
}
\makeatother

\title{\LARGE \bf
ChangeSim: Towards End-to-End Online Scene Change Detection in Industrial Indoor Environments}

\author{
	Jin-Man Park$^{1}$, 
	Jae-Hyuk Jang$^{1}$,
	Sahng-Min Yoo,
	Sun-Kyung Lee,
	Ue-Hwan Kim, \\
	and
	Jong-Hwan Kim,~\IEEEmembership{Fellow,~IEEE}
	
\thanks{The authors are with the School of Electrical Engineering, KAIST, Daejeon, 34141, Republic of Korea. \textit{e-mail: \{jmpark, jhjang, smyoo, sklee, uhkim, johkim\}@rit.kaist.ac.kr}. }
\thanks{$^{1}$These authors contributed equally to this work.}%
}

\begin{document}

\maketitle

\begin{abstract}


We present a challenging dataset, ChangeSim, aimed at online scene change detection (SCD) and more. 
The data is collected in photo-realistic simulation environments with the presence of environmental non-targeted variations, such as air turbidity and light condition changes, as well as targeted object changes in industrial indoor environments. 
By collecting data in simulations, multi-modal sensor data and precise ground truth labels are obtainable such as the RGB image, depth image, semantic segmentation, change segmentation, camera poses, and 3D reconstructions.
While the previous online SCD datasets evaluate models given well-aligned image pairs, ChangeSim also provides raw unpaired sequences that present an opportunity to develop an online SCD model in an end-to-end manner, considering both pairing and detection.
Experiments show that even the latest pair-based SCD models suffer from the bottleneck of the pairing process, and it gets worse when the environment contains the non-targeted variations. 
Our dataset is available at \url{https://sammica.github.io/ChangeSim/}.
\end{abstract}


\input{0_Intro.tex}
\input{1_DatasetFeatures}

\input{2_Method.tex}
\input{3_Experiments}

\input{4_Conclusion.tex}

\section*{Acknowledgment}
This work was supported by Institute for Information \& communications Technology Promotion (IITP) grant funded by the Korea government (MSIT) (No.2020-0-00440, Development of artificial intelligence technology that continuously improves itself as the situation changes in the real world).

\bibliographystyle{IEEEtran}

\bibliography{reference}


\end{document}

%% file: 0_Intro.tex
\section{Introduction}
\label{sec:introduction}
Scene change detection (SCD), the task of localizing changes given two scenes, past and present, is one of the most important tasks for patrol robots. SCD divides into two categories: offline and online. On the one hand, offline SCD \cite{wald2019rio,langer2020robust,fehr2017tsdf, ambrus2016unsupervised} performs two global mappings at two different time-steps; compares the two reconstructions; and detects changes between the two scenes. Since offline SCD can detect changes only after the second mapping is completed, immediate detection is impossible. On the other hand, online SCD performs frame-level detection on-the-fly during the second mapping. Therefore, it enables immediate response, which is a crucial prerequisite for surveillance. In this work, we focus on online SCD.

\begin{table*}[t]
\centering
\begin{adjustbox}{max width=\textwidth}

\begin{tabular}{l|lllllllll}
\multicolumn{7}{p{251pt}}{\footnote[2]{} Minor lighting changes with little visual variation included. }\\
\multicolumn{7}{p{251pt}}{ * Not yet publicly released.}  \\
\Xhline{2\arrayrulewidth}
\textbf{Dataset} & \textbf{Images} &  \textbf{\begin{tabular}[c]{@{}l@{}}Change\\ annotation\end{tabular}} & \textbf{Size}            & \textbf{\begin{tabular}[c]{@{}l@{}}\#change\\ class\end{tabular}} & \textbf{\begin{tabular}[c]{@{}l@{}}\#sem.\\ class\end{tabular}} & \textbf{\begin{tabular}[c]{@{}l@{}}3D\\ recon.\end{tabular}} & \textbf{Environment}       & \textbf{Target change} & \textbf{Non-target change}                                                                 \\ \hline\hline
\textit{Outdoor} & & & & & & & & &   \\
VL-CMU-CD \cite{alcantarilla2018street}    & P   & \textbf{every frame}         & 1362 images              & 1 & 9  & - & R/street-view       & structural change      & weather, light                                                                    \\
PCD \cite{sakurada2015change}              & P   & \textbf{every frame}         & 200 images               & 1 & -  & - & R/street-view       & structural change      & weather, light                                                                    \\
CD2014 \cite{wang2014cdnet}                & P   & \textbf{every frame}         & 70,000 images            & 1 & -  & - & R/street-view, cctv & dynamic object         & -                                                                                 \\
PSCD* \cite{sakurada2020weakly}             & P   & \textbf{every frame}         & 500 images               & 2 & 10 & - & R/street-view       & structural change      & weather, light                                                                    \\
CARLA-OBJCD* \cite{hamaguchi2020epipolar}   & P   & \textbf{every frame}         & 60,000 images          & 1 & 10 & - & S/street-view       & new/missing object     & -                                                                                 \\
GSV-OBJCD* \cite{hamaguchi2020epipolar}     & P   & \textbf{every frame}         & 500 images               & 1 & 10 & - & R/urban area        & new/missing object     & weather, light                                                                    \\ \hline
\textit{Indoor}  & & & & & & & & &   \\
3RScan \cite{wald2019rio}                  & \textbf{U}   & final map only & 1482 scans of 478 scenes & 1 & 7  & \checkmark & R/household         & dynamic object  & -\footnote[2]{}                                                                   \\
InteriorNet \cite{li2018interiornet}       & \textbf{U}   & final map only & Millions scans / unknown & - & 23 & \checkmark           & S/household         & dynamic object                 & light  (color, intensity)                                                          \\
Langer et al. \cite{langer2020robust}      & \textbf{U}   & final map only & 31 scans of 5 scenes     & 1 & 8  & \checkmark & R/household         & new object             & -\footnote[2]{}                                                                   \\
Fehr et al. \cite{fehr2017tsdf}            & \textbf{U}   & final map only & 23 scans of 3 scenes     & 1 & -  & \checkmark & R/household         & dynamic object         & -\footnote[2]{}                                                                   \\
Ambrus et al. \cite{ambrus2016unsupervised}& \textbf{U}   & final map only & 88 scans of 1 scenes     & 1 & -  & \checkmark           & R/office            & dynamic object         & -\footnote[2]{}                                                                   \\
\textbf{ChangeSim (ours)}                  & P\&\textbf{U}& \textbf{every frame} & \begin{tabular}[c]{@{}l@{}}80 scans of 10 scenes,\\ Approx. 130,000 images \end{tabular}  & \textbf{4} & \textbf{24} & \checkmark & S/warehouse         & \textbf{\begin{tabular}[c]{@{}l@{}}new/missing/\\ rotated/replaced object \end{tabular}}                 & \textbf{\begin{tabular}[c]{@{}l@{}}dusty air,  \\ low-illumination \end{tabular}}\\ 
\Xhline{2\arrayrulewidth}

\end{tabular}
\end{adjustbox}
\caption{Comparison of previous datasets for scene change detection. We broadly categorize these datasets as being either in \textit{outdoor} or \textit{indoor}. 
The \textit{Images} column indicates whether or not images are provided in a paired way (i.e., reference image (at time $t_0$) and query image (at time $t_1$), P: Paired, U: Unpaired). 
The \textit{Change annotation} column indicates the type of change label (\textit{every frame}: ground-truths are provided for each frame of a sequence, \textit{final map only}: ground-truths are provided for the 3D reconstruction only). 
The \textit{Environment} column indicates its domain (i.e., R: real, S: synthetic).
Note that ChangeSim is the first dataset providing both raw unpaired images and frame-by-frame change annotations, which are the prerequisite for online indoor SCD.
}
\label{table:dataset_comparison}
\end{table*}

Prior to change detection, online SCD goes through a pairing process in which a reference frame whose camera view matches that of the query (current) frame gets extracted from the past scene. At the core of the pairing process lies localization as camera poses play a key role in measuring the similarity between two frames.
In the case of online outdoor SCDs such as SCDs for street-view \cite{alcantarilla2018street} and remote sensing \cite{ji2018fully}, the availability of the global positioning system (GPS) allows robust and accurate pose estimation regardless of how much the environment has visually changed.
Therefore, most outdoor online SCD datasets assume perfect localization and provide paired image sets as input rather than pairs of raw image sequences \cite{alcantarilla2018street, sakurada2015change, wang2014cdnet, sakurada2020weakly, hamaguchi2020epipolar}.

\begin{figure}[t!]

    \includegraphics[width=8.8cm]{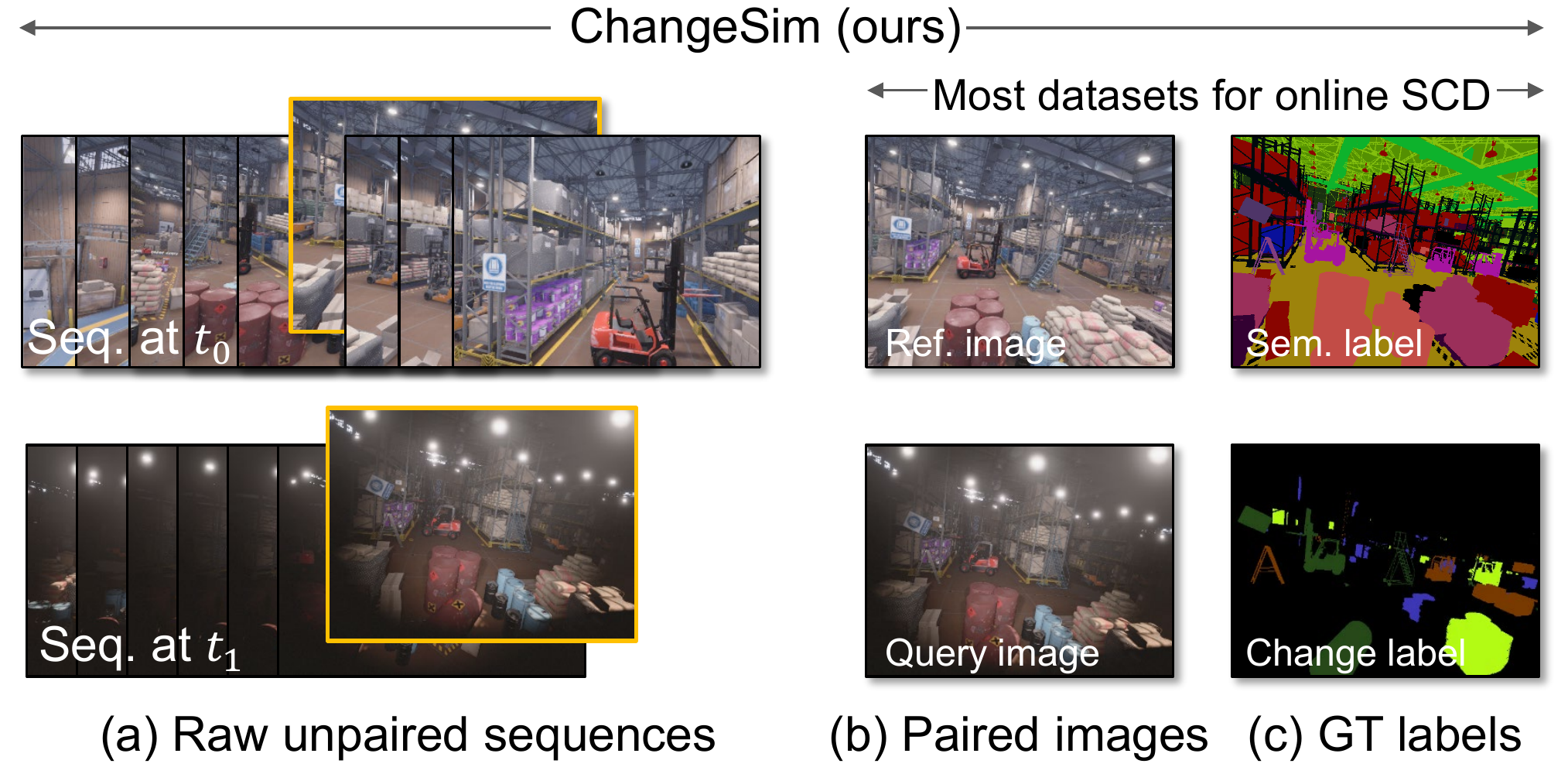}
    
    \caption{The proposed ChangeSim dataset, aimed at online scene change detection in industrial indoor environment (warehouse). While the previous online SCD datasets provided only paired images (b) and ground-truth labels (c), the proposed dataset also provides raw unpaired sequences (a) for the development of end-to-end online SCD.}
    \vspace{-15pt}
    \label{fig:main}
    
\end{figure}

On the other hand, localization becomes a challenging task for online indoor SCD since a localization algorithm can only utilize visual (+ inertial) features but not GPS signals. In addition, indoor environments frequently contain non-targeted visual changes such as air turbidity or light intensity. These variations not only degrade change detection performance for well-matched pairs, but also cause a bottleneck by deteriorating the performance of visual localization, the preceding stage of actual change detection. Due to these challenges, online indoor SCD has not displayed much progress over the past decades despite its substantial significance. Furthermore, the lack of a corresponding dataset prevents the employment of recent advances in deep learning. 
Though a set of indoor SCD datasets was reported in the literature \cite{wald2019rio,langer2020robust,fehr2017tsdf,ambrus2016unsupervised,li2018interiornet}, they merely assume the offline setting---which is not realistic in practical deployment.

We claim the following properties as an effective online indoor SCD dataset:
\begin{itemize}[]
    	\item \textit{Unpaired images}: The dataset should provide raw unpaired image sequences rather than paired image sets, since localization takes a key role in online indoor SCD.
        \item \textit{Non-targeted variation}: The dataset should include environmental non-targeted variations (e.g., light condition) because they frequently occur in real-world scenarios.
        \item \textit{Diversity}: The dataset should contain diverse types of environments and change classes. 
    \end{itemize}
The claimed properties of online indoor SCD datasets enhance the robustness and practicality of online indoor SCD models.

In this work, we introduce ChangeSim, a photo-realistic dataset towards end-to-end online indoor SCD, as shown in Figure \ref{fig:main}. To the best of our knowledge, we are the first to explicitly define the task of online indoor SCD, collect a dataset---satisfying the claimed properties---for development and evaluation of online indoor SCD; and propose baseline models. 

First of all, the proposed ChangeSim provides raw unpaired sequences captured at time $t_0$ and $t_1$ as input (for end-to-end SCD), requiring models to perform the pairing process and change detection. At the same time, ChangeSim includes paired image sets so that conventional pair-based SCD models can also benefit from ChangeSim. Next, ChangeSim provides controllable non-target variations, such as air turbidity and light condition. We show in the experiment that state-of-the-art SCD models struggle in our dataset mainly due to the bottleneck of visual localization, especially when the non-target variations are applied in the environment. Finally, ChangeSim provides rich multi-modal data and annotations, as described in Table \ref{table:dataset_comparison} (last row), including depth, semantic segmentation label, multi-class change segmentation label, 7-D pose, and 3D reconstructions.

\begin{figure}[t!]
\centering
  \includegraphics[width=8.0cm]{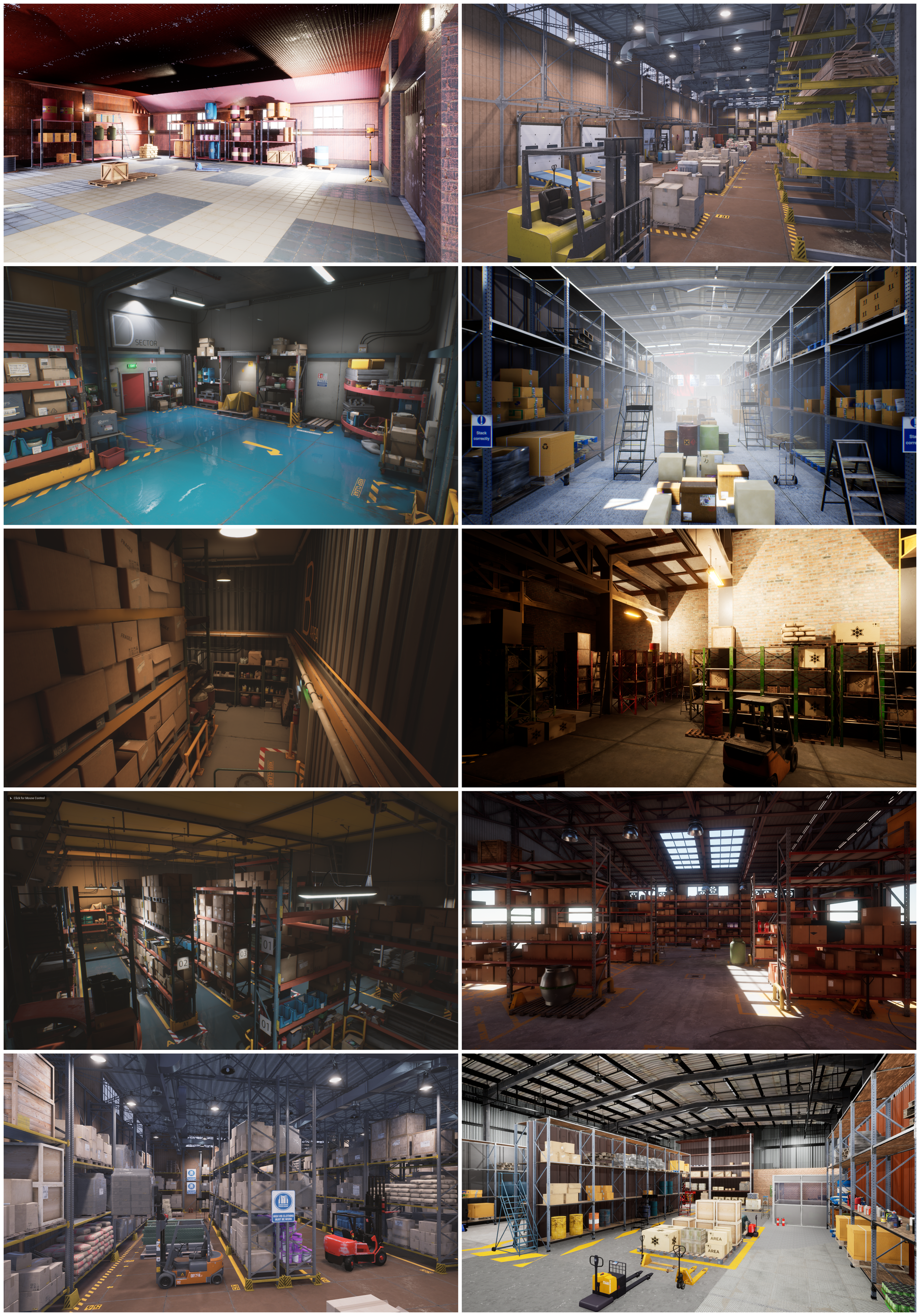}
\caption{An overview of the ChangeSim environments.}
\label{fig:overview}
\vspace{-12pt}
\end{figure}

\begin{figure*}[t]
\centering
\includegraphics[width=16cm]{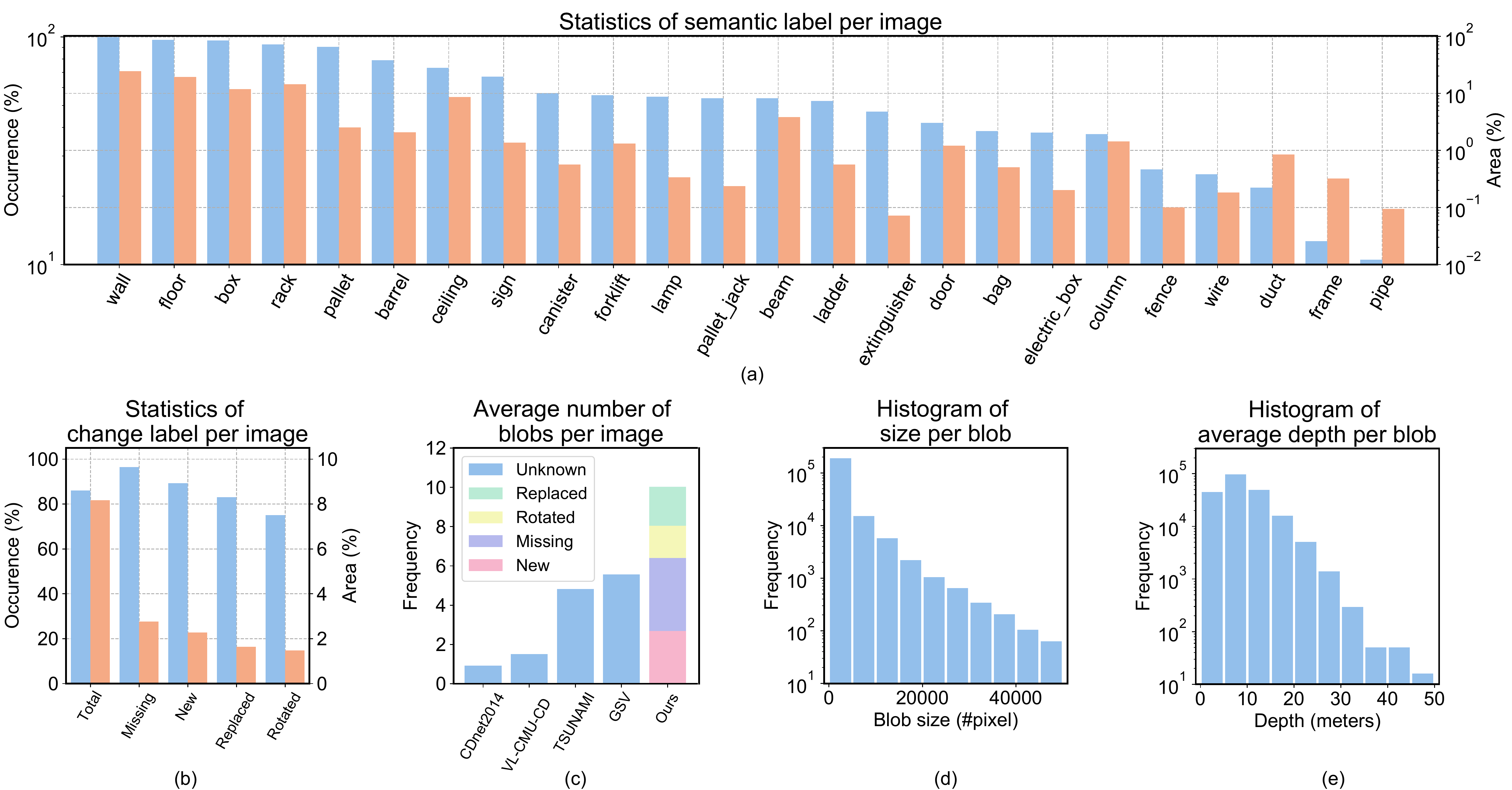}
\caption{ChangSim dataset statistics. 
(a) ChangeSim dataset semantic label statistics. (left) Average probability of occurrence per image. (right) Average area per image. 
(b) ChangeSim dataset change label statistics. (left) Average probability of appearance per image. (right) Average area per image. 
(c) Comparison of change detection datasets on blobs per image. For fair comparison, the frequency values of Panoramic image datasets (TSUNAMI, GSV) are divided by four, ensuring the same field of view (90$^{\circ}$).
(d) Histogram of blob size.
(e) Histogram of average depth per blob.
}
\label{fig:class_statistics}
\end{figure*}

In summary, the main contributions of this paper are (1) a large SCD dataset
with raw multi-modal data and rich annotations in diverse challenging
virtual indoor environments, (2) a data collection and annotation pipeline for multi-label SCD, and (3) verification of the proposed ChangeSim by evaluating a representative SLAM algorithm, semantic segmentation models, and SCD models, and (4) unveiling insights on future directions of the SCD models.


%% file: 1_DatasetFeatures.tex
\section{Dataset Features}
\label{sec:datasetfeatures}

We adopt a leading game engine, Unreal Engine 4 (UE4), to implement the virtual environment for change detection, where UE4 delivers photo-realistic environments, high-quality textures, dynamic shadows, and a variety of rendering options. 
These features not only minimize the gap between real-world and simulation environments, but also drastically lower the cost of high-quality ground-truth labels, which enables the creation of large, richly-annotated datasets. In particular, we have found that most traditional change detection datasets suffer from their sizes (i.e., lack of labeled changed objects and images) restraining learning-based methods.
We expect that the following dataset features made by photo-realistic simulation achieve better coverage of environments and scenarios.

\subsection{Diverse and Realistic Indoor Industrial Domain}
We select warehouse environments as the domain of ChangeSim. A warehouse is appropriate as the domain of change detection benchmark because it is 
1) a large-scale indoor environment compared to the household environment, 
2) an industrial environment that needs periodic surveillance, and 
3) a challenging environment for object change detection due to the densely arranged objects.
ChangeSim includes ten real-world-like warehouse environments, ranging from a small personal warehouse to a large logistics warehouse, as shown in Figure \ref{fig:overview}. 
ChangeSim contains 24 semantic categories, including domain-specific industrial objects (e.g., \textit{pallet}, \textit{forklift}, \textit{pallet\_jack}, \textit{rack}, etc.). Figure \ref{fig:class_statistics}(a) displays the list of semantic categories, and its average occurrence (\%) and area (\%) per image. It reveals the long tail-shaped, which is a natural feature of the real-world.


\subsection{Raw Data for the Whole Pipeline for Online Change Detection}
ChangeSim provides two unpaired raw sequence, reference sequence, $S_0=[f_0^1,...,f_0^i,...,f_0^n]$ and query sequence, $S_1=[f_1^1,...,f_1^j,...,f_1^m]$, captured at time $t_0$ and $t_1$, respectively, where $n$ and $m$ are the number of frames in $S_0$ and $S_1$, respectively.
Given a current frame, $f_1^j$, an online SCD model, $\textbf{\textit{D}}$, detects changes as follows:
\begin{equation}
c = \textbf{\textit{D}}(f_1^j,S_0,[f_1^0,...,f_{1}^{j-1}]),
\end{equation}
where $c \in \mathbb{R}^{w \times h \times l}$ is a multi-class change segmentation. Here, $w$, $h$, and $l$ indicate width, height, and the number of change classes, respectively.
Assuming that \textbf{\textit{D}} infer $c$ in a two-stage process, $\textbf{\textit{D}}$ can be separated into two sub-models, matching frame extractor, $\textbf{\textit{E}}$, and pair-based scene change detector, $\textbf{\textit{I}}$, as follows:
\begin{equation}
\begin{split}
f_0^i &= \textbf{\textit{E}}(f_1^j,S_0,[f_1^0,...,f_{1}^{j-1}]),\\
c &= \textbf{\textit{I}}(f_0^i,f_1^j),
\end{split}
\end{equation}
where $f_0^i$ is the extracted frame from $S_0$, where the camera view of $f_1^j$ best matches that of $f_0^i$ among frames in $S_0$.
While the traditional SCD datasets only provided the pairs, $f_1^j$ and $f_0^i$, focusing only on training and evaluation of $\textbf{\textit{I}}$, we generalize online SCD by providing the raw sequences, $S_0$ and $S_1$, so that both $\textbf{\textit{E}}$ and $\textbf{\textit{I}}$ are considered. Furthermore, an end-to-end SCD model, $\textbf{\textit{D}}$ can be developed leveraging our dataset. 

In ChangeSim, there are 80 sequences in total (20 sequences for $S_0$, 60 sequences for $S_1$), and each sequence contains 500-6,500 frames.
Each frame in $S_0$ has synchronized RGB, depth, semantic segmentation image, and camera pose. A 3D reconstruction is also provided for each reference sequence, $S_0$. 
The organization of each frame in $S_1$ is the same as that in $S_0$, except that a change segmentation label and a matching frame, $f_0^i$, are included.





\begin{table}[]
\begin{tabular}{l|ccc}
\Xhline{2\arrayrulewidth}
\textbf{Dataset}              & Langer et al. \cite{langer2020robust}& Fehr et al. \cite{fehr2017tsdf} & Ours           \\
\hline\hline
\textbf{\# of changed obj.} & 8.4 / 260          & 78.3 / 235       & \textbf{114.6} / \textbf{1146} \\
\Xhline{2\arrayrulewidth}
\end{tabular}
\centering
\caption{Average number of changed objects per environment (left), total number of changed objects in dataset (right). }
\vspace{-10pt}
\label{table:object_count}

\end{table}

\subsection{Rich Changes with Novel Multi-class Change Labels}
\subsubsection{Multi-class Change Labels}
\begin{figure}[t]
    \centering
    \subfloat{\includegraphics[width=8.2cm]{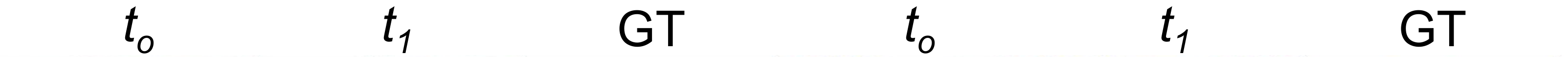}}
     \vspace{1pt}
    \subfloat{\includegraphics[width=8.2cm]{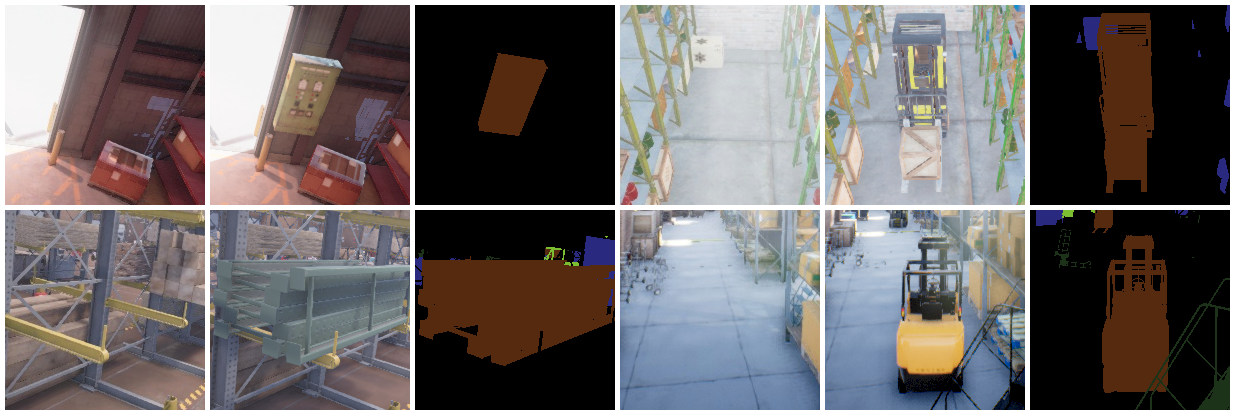}}
    \vspace{1pt}
    \subfloat{\includegraphics[width=8.2cm]{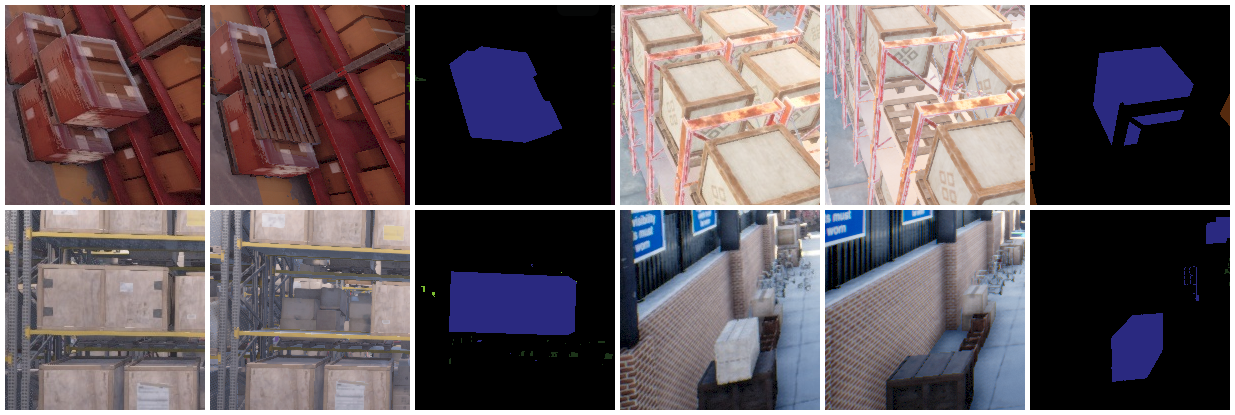}}
        \vspace{1pt}
    \subfloat{\includegraphics[width=8.2cm]{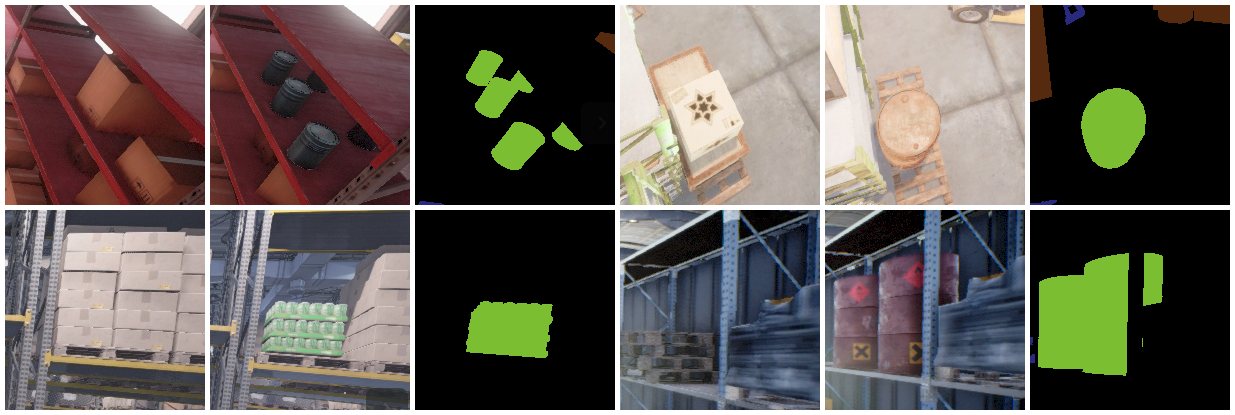}}
    \vspace{1pt}
    \subfloat{\includegraphics[width=8.2cm]{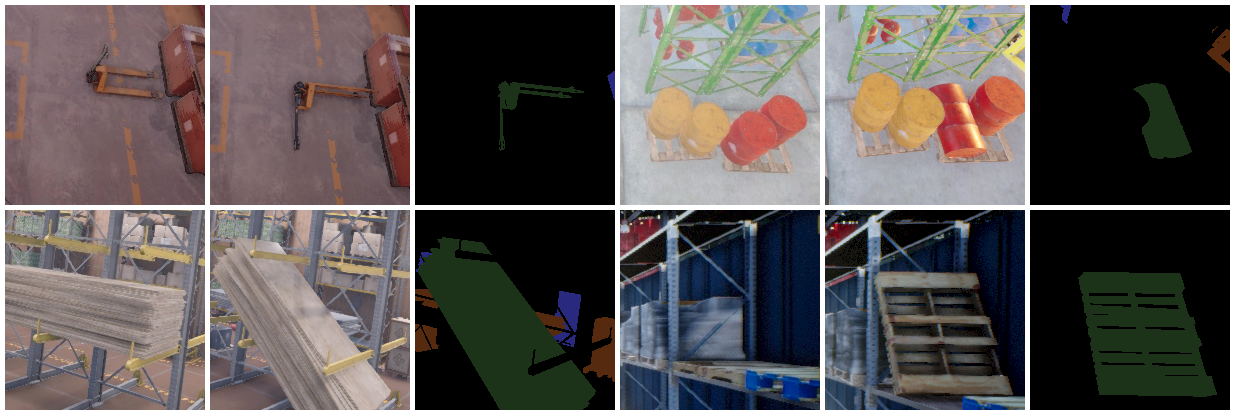}}
    
    \caption{Examples of changes in ChangeSim. Four categories of novel change types are defined, such as \textit{new} (rows 1-2), \textit{missing} (rows 3-4), \textit{replaced} (rows 5-6), and \textit{rotated} (rows 7-8), respectively.
    }
    \label{fig:change_thumbnail}
\end{figure}

While most SCD datasets provide binary change segmentation labels for the purpose of binary SCD, ChangeSim provides multi-class change segmentation labels for four change types: \textit{missing}, \textit{new}, \textit{replaced}, and \textit{rotated}, as shown in Figure \ref{fig:change_thumbnail}.
In the binary SCD, detecting pixel-level change is sufficient, but for the multi-class SCD, the ability to comprehensively consider the location, semantics, and shape of an object is required.
\subsubsection{Rich Changes}
We analyze our dataset at the levels of image and map to show that our dataset contains rich changes. At the image-level, as shown in Figure \ref{fig:class_statistics}(b-e), about 84\% of all images contain changes, and each image contains an average of 8\% changed pixels (b). Also, change blobs with fewer than 5,000 pixels are the most common (d), and the change blobs between 5 and 10 meters in depth are the most frequent (e). Besides, our dataset provides approximately 10 change blobs per image, which is the largest compared to the online SCD datasets providing image-level change labels (c).
At the map-level, as summarized in TABLE \ref{table:object_count}, we compare indoor/offline SCD datasets to our dataset according to an average number of changed objects per environment (left), and a total number of objects in the dataset (right). It shows that ChangeSim provides the most number of changed objects compared to the existing indoor SCD datasets.

\begin{figure}[ht!]
\centering
\includegraphics[width=8.2cm]{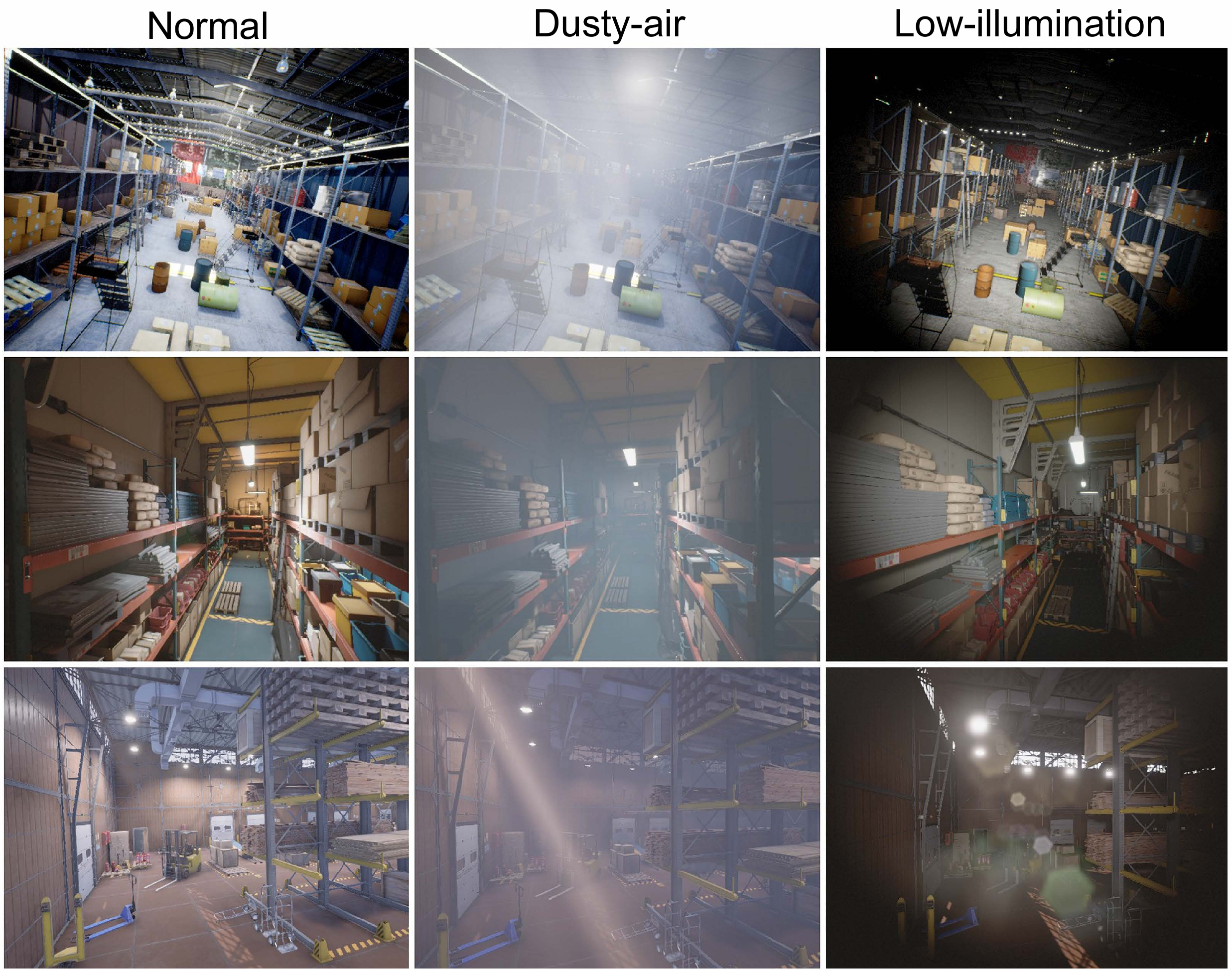}
\caption{Controllable environmental variations. }
\label{fig:challenging_scene}
\vspace{-10pt}
\end{figure}

\subsection{Controllable Environmental Variations}

\begin{figure*}[th!]
    \vspace{-5pt}

    \centering
    \includegraphics[width=18cm]{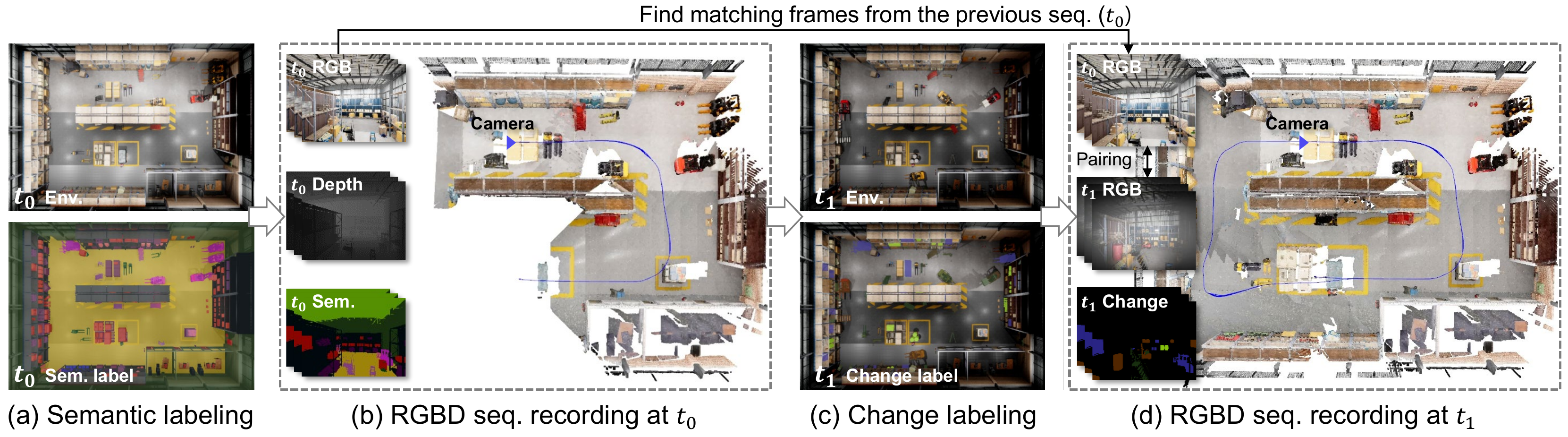}
    \caption{An overview of the data collection pipeline, including data annotation.
    In (c-d),
     \textit{new}, \textit{missing}, \textit{replaced}, and \textit{rotated} objects are marked as \textit{brown}, \textit{navy}, \textit{lime}, and \textit{green}, respectively. 
    Best viewed when magnified.
    }
    \label{fig:method}
        \vspace{-10pt}

\end{figure*}

For each reference sequence, $S_0$, ChangeSim provides three versions of the query sequence, $S_1$,: \textit{normal}, \textit{dusty-air}, and \textit{low-illumination}, allowing to directly measure the impact of environmental variations on an online SCD model. Figure \ref{fig:challenging_scene} shows examples of each scenario.
The \textit{normal} situation is when no other change has occurred other than the change of the object. 
The other two situations are devised assuming extreme cases that can occur in industrial environments.
In \textit{dusty-air}, the indoor air becomes dusty and cloudy, making it challenging to identify distant objects. Also, light is scattered as it passes through the dusty air.
\textit{Low-illumination} is a situation in which most of the indoor lighting has been removed. Instead, the camera is equipped with several flashlights. The field of view is limited to the area illuminated by the flashlights. Furthermore, the color tone becomes darker, and the texture of objects changes according to the lighting intensity.





%% file: 2_Method.tex
\section{Method}
\label{sec:Method}

To collect the raw data reflecting real-world characteristics as well as detailed annotations with minimum human efforts, we design a four-stage pipeline as shown in Figure \ref{fig:method}. 
The included stages are semantic labeling, reference sequence capturing (at time $t_0$), change labeling, and query sequence capturing (at time $t_1$).
The semantic labeling process assigns class labels to objects in the map based on their names defined by the map creators (i.e., professional designers).
Then, in the capturing process of a reference sequence, a quadrotor-type drone empowered by AirSim \cite{shah2018AirSim} moves along a hand-drawn trajectory to acquire an RGBD sequence. 
After that, we create a modified environment where some of the objects change their state, and the environment can be controlled to have optional environmental variations such as dusty-air, low illumination, and minor variation of trajectories.
Finally, in the query sequence capturing process, the drone traverses the changed environment and acquires the RGBD sequence and ground-truth labels.


 
\subsection{Semantic Labeling}
Each object in the environment has a unique name defined by professional designers, usually in the form of \textit{ModelingName\_Number} (e.g., \textit{bottle\_7}), where \textit{ModelingName} generally includes semantics and \textit{Number} is a randomly assigned integer to distinguish the same objects.
Using the form, we define various candidate names for each semantic category (e.g., candidate names for \textit{canister}: \textit{can}, \textit{bottle}, \textit{container}, etc.), and if a given object contains a candidate name, we label it with the corresponding category. 
Figure \ref{fig:method}(a) shows the environment from the top-view where the semantic labeling is completed.

\subsection{Reference Sequence Capturing}
\label{subsec:reference_sequence}
\subsubsection{Trajectory Drawing}
Auto-generated trajectories have the advantage of being able to include various movements, so they are mainly used for synthetic SLAM datasets.
However, auto-generated trajectories are often unsuitable for surveillance tasks, such as when the trajectory is unnatural or when it gets too close to an obstacle.
As ChangeSim aims at surveillance-oriented change detection, we adopt hand-drawing trajectories to minimize this side-effect.
We design the trajectories mainly in the form of traversing one lap through the corridors, keeping the  distance to nearby obstacles greater than a pre-defined threshold to obtain a view suitable for change detection.

\subsubsection{RGBD-SLAM based 3D Mapping}
The multirotor traverses the environment and collects RGBD, 7-D pose, semantic labels for each frame, and 3D reconstruction for each sequence along the hand-drawn trajectories (see Figure \ref{fig:method}(b)). 
One of the most latest RGBD-SLAM models, RTABMAP \cite{labbe2019rtab}, is employed to collect pose and 3D reconstruction.
The data is collected at a velocity of up to 3 m/s, an acquisition rate of 1-3 fps, and a clock-speed of 0.03. 
Each sequence took about 30 minutes to 1 hour.

\subsection{Change Labeling}
We limit the candidate semantic classes to be changed to 16 classes out of 24 in total, mainly \textit{things} (i.e., countable objects), because \textit{stuff} objects (i.e., uncountable objects, such as \textit{floor}, \textit{ceiling}, etc.) have little probability of changing within a short time in the real world.
Figure \ref{fig:method}(c) shows the result of applying the object changes with the change labels.


\textit{\textbf{Missing}}: One of the advantages of virtual environments, including the Unreal Engine, is that the types of rendering are selectable. We disable RGB rendering and Depth rendering for the selected object.
Through this, the object becomes transparent, that is, it does not appear in the RGB and Depth images, but is still displayed in the segmentation image, resulting in \textit{missing} type of change label. \textit{\textbf{New}}: We randomly select 3D modelings and place them on the floor, shelves, or other objects in the environment.
\textit{\textbf{Replaced}}: After deleting the selected object, another 3D modeling is placed in the same place. The replaced one may or may not belong to the same semantic class as the deleted object.
\textit{\textbf{Rotated}}: The selected object is rotated randomly in the x, y, and z axes.


\subsection{Query Sequence Capturing}
\subsubsection{Data acquisition in a changed environment} 
As shown in Figure \ref{fig:method}(d), the multirotor traverses the changed environment and collects RGB, Depth, estimated semantic \& change labels, and 7-D poses. The poses are obtained employing RTABMAP's localization mode, where the SLAM algorithm stops mapping and does only localization (i.e., estimate 7-D poses), given a 3D reconstruction and its corresponding RGBD sequence. Note that the multirotor is localized in the 3D map reconstructed from the traverse at time $t_0$, as shown in (d).
At this time, the multirotor follows the noise-added version of the trajectory used in the previous mapping (b), and Gaussian noise $N(0,0.3^2)$ is applied to each coordinate of waypoints in the trajectory.

We capture three scenarios for each sequence: normal, dusty-air, low-illumination.
In the case of \textit{Dusty-air}, the weather control option provided by AirSim is used (Dust 0.5, Foggy 0.5). 
For the \textit{low-illumination}, almost all lights in the environment are turned off, or their brightness is minimized. Instead, two forward-facing lights are installed at the front of the multirotor, ensuring a minimum amount of sight.


\subsubsection{Matching frame extraction}
For the current frame, a frame with the closest L1 distance from the reference sequence is selected using 7-D poses, and then the two frames are paired.
While this method is the most intuitive and straightforward, it highly depends on the visual SLAM algorithm's localization performance and is therefore vulnerable to visual changes.
As the pairing becomes inaccurate, the overlapping area within a pair also reduces, which leads to deterioration in change detection.


%% file: 3_Experiments.tex
\section{Baseline and analysis}
\label{sec:experimental_evaluation}

In order to verify the usefulness of ChangeSim, we propose a straightforward baseline model of a two-step strategy for online SCD and evaluate each module of this baseline: 1) matching frame extraction and 2) pair-based scene change detection.
In matching frame extraction, a pairing process is performed using the estimated pose obtained through the SLAM algorithm. In pair-based scene change detection, changes are detected using the latest learning-based SCD models, CSCDNet \cite{sakurada2020weakly} and ChangeNet \cite{varghese2018changenet}.

In particular, we measure the impact of environmental variation for each module and show that the existing models are vulnerable to environmental variation.
Also, in 2), the effect of multi-class on SCD is measured.
While existing models manage to identify changes, they show weakness in classifying the type of change, implying that they perform as a pixel-level change detector function rather than perceiving the essence of changes.
As an example of using semantic labels, we show the effect of simple fine-tuning-based transfer learning on the online SCD.
We hope that various approaches to online SCD will be developed and evaluated without restraint via ChangeSim, towards end-to-end, multi-class, and visually robust online SCD.


\subsection{Train-test Split}
To establish concrete benchmarks, we split our data into training and test sets according to the provided maps. We selected 6 maps (12 sequences, 2 sequences for each map) as the training set (rows 1-3, Figure \ref{fig:overview}) and the remaining 4 maps (8 sequences, 2 sequences for each map) as the test set (rows 4-5). 

\subsection{Matching Frame Extraction}
\begin{figure}[t!]
\centering
\includegraphics[width=8cm]{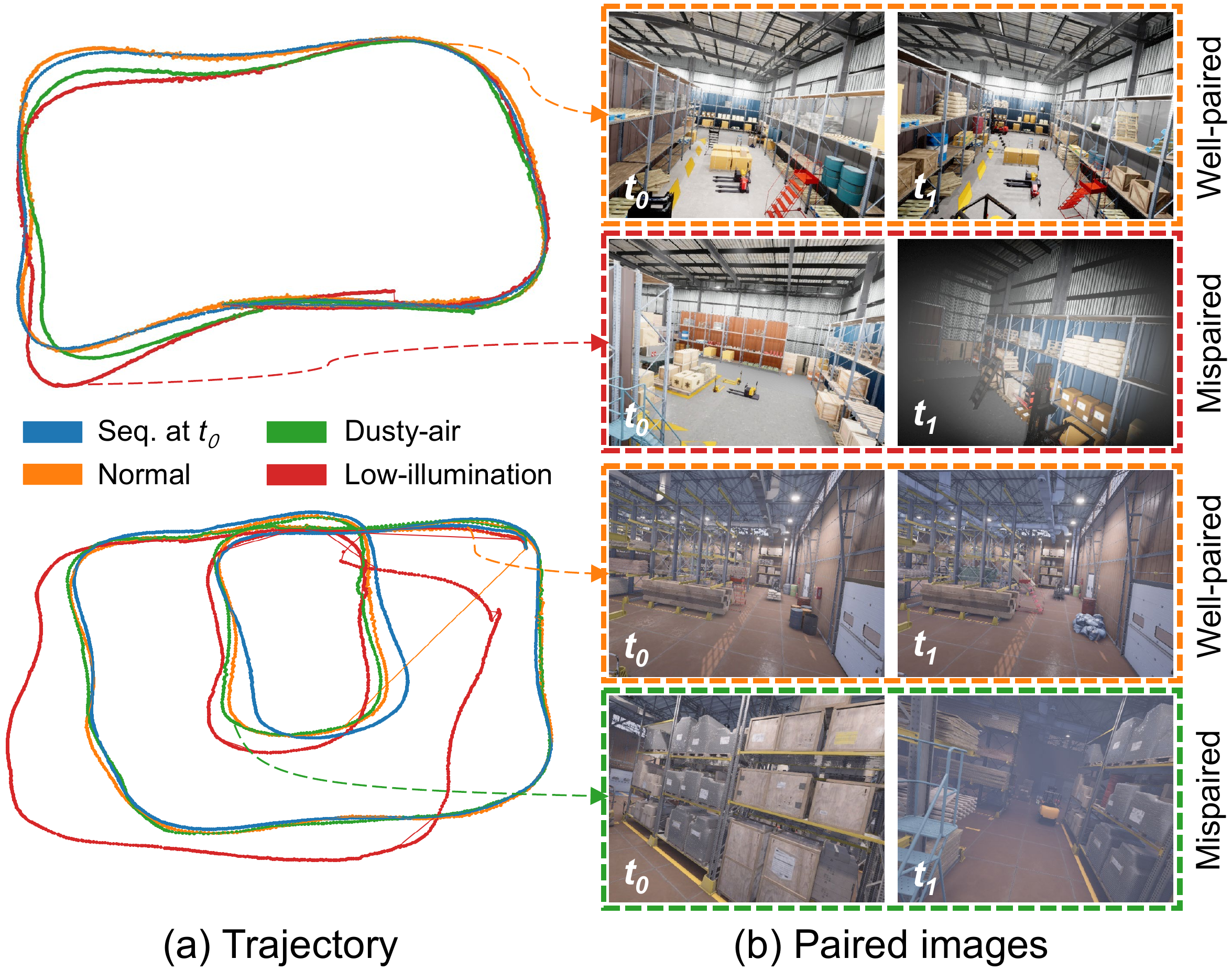}
\caption{Qualitative results matching frame extraction for the two reference sequences (blue). (a) The trajectories from the query sequences drawn as orange, green and red, respectively. (b) Pairing results are shown.
}
\label{fig:trajectory}
\end{figure}

Given the current frame, visual localization is performed estimating the 7-D pose through RTABMAP, and the frame with the least L1 distance is extracted from the reference sequence, $S_0$.
Table \ref{table:trajectory} shows visual localization performance measured by absolute trajectory error (ATE), according to three environmental variations for six randomly selected sequences.
Assuming that the degree of environmental variation becomes more severe in the order of \textit{normal}$<$\textit{dusty-air}$<$\textit{low-illumination}, the visual localization tend to become more and more inaccurate as the environmental variation becomes more severe.
Figure \ref{fig:trajectory}(a) visualizes the estimated trajectory for each of the three environmental variation scenarios for the two reference sequences, and examples of paired images are shown in Figure \ref{fig:trajectory}(b). 
The results confirm that the performance degradation of visual localization is directly linked to low pairing quality, which severely interferes with the following pair-based SCD.


\begin{table}[]
\begin{adjustbox}{max width=\columnwidth}
\begin{tabular}{l|cccccc}
\Xhline{2\arrayrulewidth}
                 &\multicolumn{6}{c}{ATE (m)}         \\
                 & Seq. 1        & Seq. 2        & Seq. 3        & Seq. 4        & Seq. 5        & Seq. 6        \\
                 
\hline\hline
\textit{Normal}           & 0.46          & 0.83          & 1.11          & 1.24          & 1.33          & 1.52          \\
\textit{Dusty-air  }      & \textbf{4.43}          & 0.99          & 1.29          & 1.34          & 1.31          & 1.69          \\
\textit{Low-illumination} & 2.66 & \textbf{1.11} & \textbf{1.32} & \textbf{3.37} & \textbf{4.04} & \textbf{1.93} \\
\Xhline{2\arrayrulewidth}
\end{tabular}
\end{adjustbox}
\caption{Impact of environmental variation on visual localization. Absolute trajectory errors (ATEs) were shown for the six randomly chosen query sequences.} 
\label{table:trajectory}
\end{table}

\begin{figure*}[t]
    \centering
    \subfloat{\includegraphics[width=17.4cm]{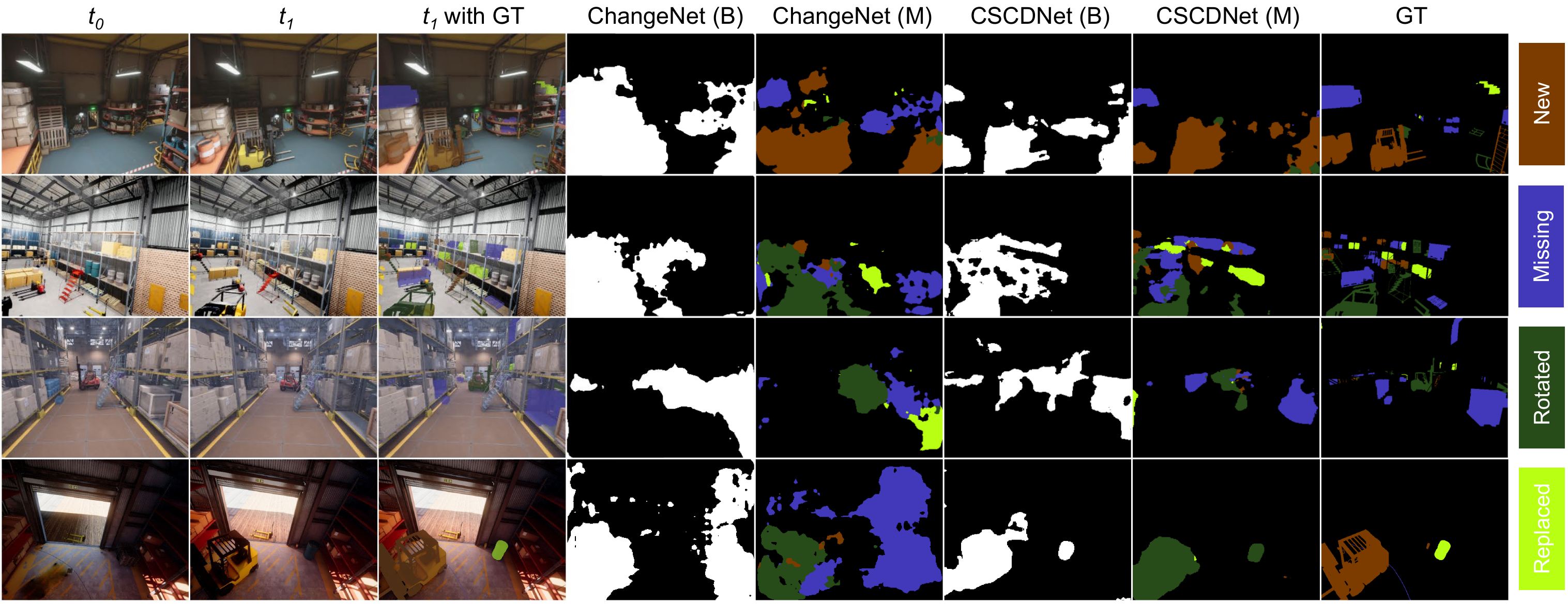}}
    \caption{Qualitative comparison of the two representative pair-based SCD models, ChangeNet \cite{varghese2018changenet} and CSCDNet \cite{sakurada2020weakly} for the ChangeSim dataset. The RGB images at $t_0$ (column 1) are obtained by the matching frame extraction process. (B) and (M) indicate binary SCD and multi-class SCD, respectively. 
    }
    \label{fig:ChangeDetectionResult}
    \vspace{-10pt}
\end{figure*}

\subsection{Pair-based Scene Change Detection}

Given a pair of the reference and query images, SCD is performed to infer changes for the present view. We benchmarked the following representative pair-based SCD models: ChangeNet \cite{varghese2018changenet} which consists of siamese networks \cite{mueller2016siamese} and fully convolutional networks (FCNs), and CSCDNet \cite{sakurada2020weakly} which consists of siamese networks and correlation networks \cite{dosovitskiy2015flownet}.

For the implementation details, we used the Adam optimizer with the learning rate of $1\times10^{-4}$. 
The ImageNet-pretrained ResNet backbone was used, unless otherwise specified.
We used the batch size of $16$ and minimized Cross-entropy loss for 20 epochs. Horizontal flips and color jittering were applied during training. The training time took approximately 24 hours with a single RTX2080Ti. 
The inputs to the models are a pair of RGB images, where the images are resized to 240$\times$320, and 256$\times$256 for the ChangeNet and CSCDNet, respectively, to meet their input-size requirements. In particular, the pairs are obtained by the matching frame extraction; thus, both train/test sets contain mismatched pairs and well-matched pairs. We report macro-F1 (i.e., average of per-class F1) and mean intersection over union (mIoU) for the following SCD results. The results are from multi-class SCDs unless otherwise specified.

\begin{table}[t]
\centering
\begin{adjustbox}{max width=\columnwidth}
\begin{tabular}{l|ccc|cccccc}
\Xhline{2\arrayrulewidth}
\multirow{2}{*}{}              & \multicolumn{3}{c|}{Binary} & \multicolumn{6}{c}{Multi-class} \\ 
                              & C    &S & mIoU    & M        & N      & Re      & Ro       & S    & mIoU  \\ \hline\hline
\multicolumn{1}{c|}{ChangeNet} & 17.6 & 73.3 & 45.4   & 6.9   & 9.1   & 6.6    & 11.6     & 80.6  &  23.0      \\
\multicolumn{1}{c|}{CSCDNet}   & 22.9 & 87.3 & 55.1   & 6.0   & 12.4  & 7.9    & 17.5     & 90.2  &  26.8      \\
\Xhline{2\arrayrulewidth}
\end{tabular}
\end{adjustbox}
\caption{Impact of multi-class learning. IoUs of Binary and Multi-class SCD are reported (C: \textit{change}, S: \textit{static}, M: \textit{missing}, N: \textit{new}, Re: \textit{replaced}, Ro: \textit{rotated}).}
\label{table:multiclass}
\end{table}


\begin{table}[]
\centering
\begin{adjustbox}{max width=\columnwidth}
\begin{tabular}{l|cccccc}
\Xhline{2\arrayrulewidth}
          & \multicolumn{2}{c}{\textit{Normal}} & \multicolumn{2}{c}{\textit{Dusty-air}} & \multicolumn{2}{c}{\textit{Low-illumination}} \\
          & mIoU         & macro-F1     & mIoU           & macro-F1      & mIoU              & macro-F1             \\ \hline\hline
ChangeNet & 23.0         & 29.8         & 21.3           & 26.0          & 20.9              & 24.6              \\
CSCDNet   & 26.8         & 30.6         & 24.4           & 27.2          & 22.6              & 24.5             \\ 
\Xhline{2\arrayrulewidth}
\end{tabular}
\end{adjustbox} 
\caption{Impact of environmental variations. mIoUs and macro-F1s are reported for the three cases of environmental variations.}
\label{table:env_change}
\vspace{-10pt}
\end{table}

\begin{table}[t]
\centering
\begin{adjustbox}{max width=\columnwidth}
\begin{tabular}{l|ccccc}
\Xhline{2\arrayrulewidth}
           &  \multicolumn{3}{c}{ChangeSim} & Pascal VOC & Cityscapes \\
           & mIoU  & fwIoU & P. Acc. ($\%$) & mIoU & mIoU \\ \hline\hline
           
DeepLab-V2 & 34.65 & 58.69 & 73.48 & 79.7 &  70.4 \\
DeepLab-V3 & 39.80 & 62.95 & 76.68 & 85.7 &  81.3 \\

\Xhline{2\arrayrulewidth}   
\end{tabular}
\end{adjustbox}
\caption{Semantic segmentation results for ChangeSim. For an intuitive comparison, the mIoU performances for PASCAL VOC \cite{everingham2015pascal} and CityScapes \cite{cordts2016cityscapes} are also reported.}
\vspace{-7pt}
\label{table:seg_result}

\end{table}

\begin{table}[t]
\centering
\begin{adjustbox}{max width=\columnwidth}
\begin{tabular}{c|cccccc}
\Xhline{2\arrayrulewidth}
          & \multicolumn{2}{c}{Scratch} & \multicolumn{2}{c}{ImageNet} & \multicolumn{2}{c}{ChangeSim} \\ 
          & mIoU     & macro-F1     & mIoU       & macro-F1     & mIoU       & macro-F1           \\ \hline\hline
ChangeNet & 18.8      & 22.0          & 21.1        & 25.0          & 23.0       & 29.8               \\ 
CSCDNet   & 22.1      & 25.3          & 26.1        & 30.2          & 26.8       & 30.6               \\ 

\Xhline{2\arrayrulewidth}
\end{tabular}
\end{adjustbox}
\caption{Impact of pre-training. SCD results are shown for the three pre-trained backbone ResNet.} 
\label{table:pretraining}
\vspace{-10pt}
\end{table}

\subsubsection{Impact of Multi-class Learning}
In Table \ref{table:multiclass}, the effect of multi-class learning on SCD is reported. There is a significant performance drop when performing multi-class SCD, implying that the models' abilities to distinguish characteristics of each change category (missing, new, replaced, or rotated) are poor while they are good at detecting pixel-level changes.
This is because multi-class SCD requires a comprehensive understanding of changes in semantics, position, size, and pose of an object to distinguish change categories.

\subsubsection{Impact of Environmental Variations}
Table \ref{table:env_change} shows the impact of environmental variations between the two sequences, $S_0$ and $S_1$, on the model performance. Note that environmental variations are not included in the training set. 
As expected, it is confirmed that model performances decrease as the environmental variations become more severe.
In particular, the performance significantly decreases in \textit{low-illumination}, which seems to be due to the incorrect pairing process. That is, two frames of a pair barely overlap their camera view.

\subsubsection{Impact of Pre-training}
We performed a simple fine-tuning-based transfer-learning experiment to confirm the effectiveness of ChangeSim's semantic labels for SCD.
As an upstream task, we trained representative semantic segmentation models, DeepLab-V2 \cite{chen2017deeplab} and DeepLab-V3 \cite{chen2018encoder}, using semantic segmentation labels.
The original image size of 640$\times$480 was used, and the cross-entropy loss was minimized using SGD, with the batch size of 3.
Each model was trained up to 20 epochs without any augmentation.
Table \ref{table:seg_result} shows the evaluation results in the test set.
The performance difference of mIoU between ChangeSim and other datasets, PASCAL VOC and Cityscapes, is analyzed to be due to a large number of classes and the appearance characteristics of industrial indoor environment objects, in particular, the presence of sharp and complex objects such as wires and racks. 

For the downstream task, we took the trained backbone ResNet50 of DeepLab-V3 as the backbone of ChangeNet and CSCDNet, respectively.
The remaining implementation details were set the same as in other experiments. 
As shown in Table \ref{table:pretraining}, the SCD performance when using the backbone trained from ChangeSim was higher than that trained from ImageNet, implying that semantic-awareness helps SCD.


\subsection{Qualitative Evaluation}

\subsubsection{Scene Change Detection}
Figure \ref{fig:ChangeDetectionResult} visualizes the SCD results of two pair-based SCD models, CSCDNet and ChangeNet. 
Binary SCD and multi-class SCD are denoted by B and M, respectively. Consistent with the results shown in Table \ref{table:multiclass}, CSCDNet caught the changes better than ChangeNet.
In the multi-class SCD setting, the silhouette of changed objects is better captured compared to the binary SCD. However, the changes are often misclassified.

\subsubsection{Semantic Segmentation}

We also show the qualitative results for the segmentation networks, as shown in Figure \ref{fig:seg}. The results show that the segmentation network trained on the train split of ChangeSim works well on both a) the test split of ChangeSim and b) real-word samples, proving that ChangeSim successfully represents the real-world characteristics.

%% file: 4_Conclusion.tex
\section{Conclusion}
\label{sec:conclusion}

We have presented \textit{ChangeSim}, a new dataset for online indoor scene change detection (SCD), satisfying the following properties for the effective development and evaluation of online SCDs: unpaired image sets; non-targeted variation; and diversity.
We have verified the applicability of ChangeSim through a two-stage online SCD model, whose stages are named as 1) matching frame extraction; and 2) pair-based SCD. 
In particular, we have found that existing pair-based SCD models suffer from the bottleneck of 1), mainly due to the vulnerability of visual localization. 
This result implies that an end-to-end and visually robust method should be developed for the application of online indoor SCD models in practice.
We hope that ChangeSim will complement other datasets and benchmarks, and stimulate the development of the online indoor SCD, leveraging the claimed properties.

\begin{figure}[t!]
\centering
\includegraphics[width=8.2cm]{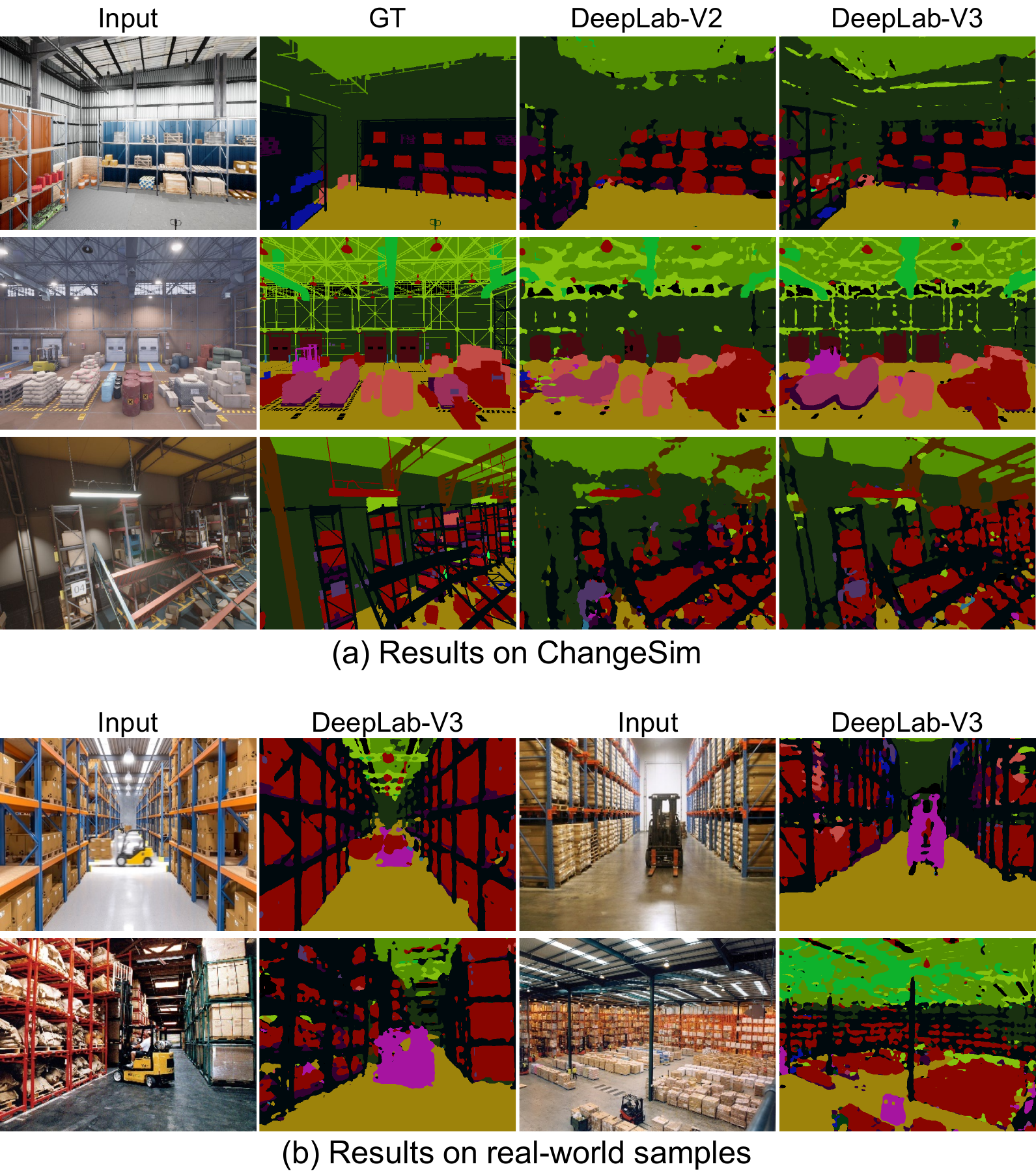}
\caption{Qualitative test results of the representative semantic segmentation networks.}
\label{fig:seg}
\vspace{-10pt}
\end{figure}